\documentclass[12pt]{article}
\usepackage{graphicx}
\usepackage{booktabs}
\usepackage{hyperref}
\usepackage{natbib}
\usepackage{geometry}
\usepackage{setspace}
\usepackage{array}
\usepackage{multirow}
\usepackage{float}
\usepackage{longtable}
\usepackage{url}
\title{\textbf{Prompt Sensitivity and Answer Consistency in Small Open-Source Language Models for Clinical Question Answering: An Empirical Evaluation for Low-Resource Healthcare Deployment}}

\author{Shravani Hariprasad\\
\textit{Independent Researcher}}

\date{}

\begin{document}

\maketitle

\begin{abstract}

\textbf{Background:} Artificial intelligence is increasingly 
deployed in healthcare workflows, and small open-source language 
models are gaining attention as viable tools for low-resource 
settings where cloud infrastructure is unavailable. Despite 
their growing accessibility, the reliability of these models, 
particularly the stability of their outputs under different 
phrasings of the same clinical question, remains poorly 
understood.

\textbf{Objective:} This study systematically evaluates prompt 
sensitivity and answer consistency in small open-source language 
models on clinical question answering benchmarks, with 
implications for low-resource healthcare deployment.

\textbf{Methods:} Five open-source language models spanning
distinct architectural and training paradigms were evaluated
across three clinical question answering datasets (MedQA,
MedMCQA, PubMedQA) using five controlled prompt style
variations, yielding 15,000 total inference calls conducted
locally on consumer CPU hardware without fine-tuning.
Consistency scores, accuracy, and instruction-following
failure rates were measured and interpreted in the context
of each model's architectural design.

\textbf{Results:} Consistency and accuracy were largely
independent across models and datasets. Gemma 2 achieved
the highest consistency scores (0.845--0.888) but the
lowest accuracy (33.0--43.5\%), producing perfectly
consistent yet incorrect answers on 77 of 200 MedQA
questions (38.5\%), a failure mode termed \textit{reliable
incorrectness}. Llama 3.2 demonstrated moderate consistency
(0.774--0.807) alongside the highest accuracy (49.0--65.0\%).
Roleplay prompts consistently reduced accuracy across all
models and datasets, with Phi-3 Mini showing the largest
decline of 21.5 percentage points on MedQA. Instruction-following
failure rates varied by model and were not determined by
parameter count, with Phi-3 Mini exhibiting the highest
UNKNOWN rate at 10.5\% on MedQA. Meditron-7B, a
domain-pretrained model without instruction tuning, exhibited
near-complete instruction-following failure on PubMedQA
(99.0\% UNKNOWN rate), demonstrating that domain knowledge
alone is insufficient for structured clinical question
answering.

\textbf{Conclusions:} High consistency does not imply 
correctness in small clinical language models; models can 
be reliably incorrect, representing a potentially dangerous 
failure mode in clinical decision support. Roleplay prompt 
styles should be avoided in healthcare AI applications. 
Among the models evaluated, Llama 3.2 demonstrated the 
strongest balance of accuracy and reliability for 
low-resource deployment. These findings highlight the 
necessity of multidimensional evaluation frameworks that 
assess consistency, accuracy, and instruction adherence 
jointly for safe clinical AI deployment.

\end{abstract}


\newpage
\tableofcontents
\newpage

\section{Introduction}

Artificial intelligence is increasingly being integrated into 
healthcare workflows, including clinical decision support, 
medical documentation, and exam-style clinical question answering. 
While much attention has focused on large cloud-based models, 
smaller open-source language models are gaining importance 
because they can run locally on standard CPUs, making them 
viable for rural clinics, community hospitals, and low-resource 
health systems with limited infrastructure \citep{garg2024slm}. 
As these smaller models become more accessible for real-world 
deployment, questions about their reliability become critical. 
In clinical settings, even small inaccuracies can have 
significant consequences, and the safety of AI systems depends 
not only on their overall performance but also on how stable 
and predictable their outputs are under different conditions. 
Despite increasing adoption, the reliability of small, locally 
deployable models in healthcare remains insufficiently studied.

Most existing evaluations of clinical language models focus 
primarily on accuracy, that is, whether a model selects the 
correct answer on a given benchmark \citep{bedi2025systematic}. 
However, accuracy alone does not capture how stable a model's 
outputs are when the same clinical question is phrased 
differently. In safety-critical domains such as healthcare, 
this distinction is important. A model that gives different 
answers depending on how a question is worded cannot be 
considered reliable, regardless of its aggregate accuracy score.

Prior work has not systematically evaluated whether small, 
CPU-runnable language models provide consistent answers under 
controlled prompt variation in clinical contexts. The CLEVER 
framework \citep{cleverjmir2025} evaluates clinical LLM outputs 
through physician preference but does not assess response 
stability across prompt reformulations. Ngweta et al. 
\citep{ngweta2024robustness} quantify prompt brittleness in 
general NLP tasks but do not examine clinical datasets or 
safety implications. Kim et al. \citep{kim2025hallucination} 
demonstrate that foundation models frequently produce confident 
but incorrect medical outputs, yet do not empirically measure 
variability under prompt changes. Surveys on small language 
models in healthcare \citep{garg2024slm} discuss efficiency 
and deployment tradeoffs but do not analyze answer consistency 
as a reliability dimension. The present study addresses this gap.

This study addresses three research questions:

RQ1: Do small open-source language models produce consistent 
answers when semantically equivalent clinical questions are 
phrased differently?

RQ2: Is answer consistency correlated with clinical question 
answering accuracy?

RQ3: How do different prompt styles influence the reliability 
of model outputs?

To investigate these questions, this work evaluates five 
open-source language models—Phi-3 Mini (3.8B), Llama 3.2 (3B), 
Gemma 2 (2B), Mistral 7B, and Meditron-7B (domain-pretrained)—
across three established clinical question answering benchmarks 
(MedQA, MedMCQA, and PubMedQA). Each question is evaluated under 
five controlled prompt styles designed to simulate realistic 
query reformulation. A quantitative consistency score is 
introduced to measure how often a model produces the same 
answer across prompt variations for the same question, and 
this metric is analyzed alongside overall accuracy, per-style 
accuracy, and instruction-following failure rates. All 
experiments were conducted locally on consumer CPU hardware 
using unmodified base models without medical fine-tuning, 
reflecting realistic deployment conditions in low-resource 
healthcare environments.

The analysis yields several key findings. First, consistency 
and accuracy emerge as independent performance dimensions: 
models that produce highly stable answers across prompt 
variations are not necessarily more accurate. Second, roleplay 
prompt styles consistently reduce performance across models 
and datasets, indicating that persona-based prompting may 
degrade reliability in clinical question answering tasks. 
Third, instruction-following failures vary independently of 
model size, suggesting that larger parameter counts do not 
guarantee more dependable outputs. Finally, Meditron-7B, a 
domain-pretrained model without instruction tuning, exhibits 
near-complete instruction-following failure despite encoding 
substantial medical knowledge, demonstrating that domain 
knowledge and instruction-following capability are distinct 
requirements for clinical AI deployment.

The remainder of this paper is structured as follows. 
Section~\ref{sec:related} reviews related work on clinical 
LLM evaluation, prompt sensitivity, and small language models 
in healthcare. Section~\ref{sec:methods} describes the 
datasets, models, prompt variation design, inference setup, 
and evaluation metrics. Section~\ref{sec:results} presents 
experimental results. Section~\ref{sec:discussion} discusses 
clinical implications, limitations, and future directions. 
Section~\ref{sec:conclusion} concludes the paper.

\section{Related Work}\label{sec:related}

Most prior work evaluating large language models in healthcare 
has focused on accuracy, reasoning quality, and output safety. 
The CLEVER framework \citep{cleverjmir2025} evaluates clinical 
LLM outputs through blinded physician preference on tasks 
including summarization and question answering, demonstrating 
that smaller domain-specific models can outperform larger 
general-purpose systems on clinically relevant criteria. 
Kim et al. \citep{kim2025hallucination} systematically examine 
medical hallucinations in foundation models, highlighting the 
dangers of confident but incorrect outputs and their potential 
to mislead clinical decision-making. While these studies 
provide critical insights into output quality and safety, 
neither investigates whether models produce consistent answers 
when the same clinical question is phrased differently, 
leaving prompt sensitivity largely unexplored in healthcare AI.

Growing interest in small, CPU-runnable language models has 
emerged in response to the practical constraints of healthcare 
deployment in low-resource settings. Garg et al. 
\citep{garg2024slm} survey small language models in healthcare, 
documenting their efficiency advantages and suitability for 
local deployment under limited infrastructure. The MedQA 
\citep{jin2021disease}, MedMCQA \citep{pal2022medmcqa}, and 
PubMedQA \citep{jin2019pubmedqa} benchmarks provide standardized 
evaluation tasks across diverse clinical question formats and 
have been widely adopted for assessing compact models. These 
works collectively advance the case for small model deployment 
in accessible healthcare AI but focus primarily on task 
accuracy and computational efficiency rather than answer 
stability under prompt variation.

Prompt sensitivity, defined as the tendency of model outputs 
to change in response to semantically equivalent but 
stylistically different inputs, has been studied in general 
NLP research. Ngweta et al. \citep{ngweta2024robustness} 
demonstrate that minor changes in prompt formatting cause 
measurable performance fluctuations across multiple models 
and benchmarks and propose mitigation strategies to reduce 
this brittleness. However, such investigations have not been 
extended to clinical question answering, and the safety 
implications of prompt sensitivity for healthcare AI 
deployment remain largely unexplored.

Taken together, these lines of research motivate the present 
study, which systematically evaluates prompt sensitivity and 
answer consistency in small open-source clinical language 
models. This work complements prior studies on accuracy, 
hallucination, and general prompt robustness by focusing on 
reliability under realistic deployment constraints.

\section{Methods}\label{sec:methods}

This study follows a controlled empirical evaluation design to assess the reliability of small language models under prompt variation in clinical question answering tasks.

\subsection{Datasets}

Model consistency and accuracy were evaluated using three established 
clinical question answering benchmarks. MedQA \citep{jin2021disease} 
contains multiple-choice questions derived from the United States Medical 
Licensing Examination (USMLE), representing clinical knowledge required 
for medical licensure. MedMCQA \citep{pal2022medmcqa} consists of multiple-choice 
questions from Indian medical entrance examinations (AIIMS/NEET), 
covering diverse medical specialties including pharmacology, 
anatomy, and pathology. Although the dataset includes subject labels, 
the evaluation was conducted on a randomly sampled subset of 200 
questions without stratification by specialty. PubMedQA 
\citep{jin2019pubmedqa} presents biomedical research questions paired 
with PubMed abstracts, requiring yes, no, or maybe answers based on 
provided evidence rather than memorized knowledge.

From each dataset, 200 questions were randomly sampled using a fixed random seed 
(seed = 42) to ensure reproducibility. This sample size was selected to balance 
statistical coverage with the computational constraints of local CPU inference 
across multiple models and prompt variations, resulting in 15,000 total inference 
calls.

\subsection{Models}

Five open-source language models spanning 2B--7B parameters were
evaluated, including one domain-adapted medical pretraining model.
Table~\ref{tab:model_arch} summarises the architectural and training
characteristics of each model. The four instruction-tuned models
differ in their fine-tuning methodology, pretraining scale, and
prompt template conventions in ways that plausibly influence prompt
sensitivity behaviour.

Phi-3 Mini \citep{abdin2024phi3} (3.8B, Microsoft) is a dense
decoder-only transformer fine-tuned using Supervised Fine-Tuning
(SFT), Direct Preference Optimisation (DPO), and Reinforcement
Learning from Human Feedback (RLHF). Its pretraining relies heavily
on high-quality synthetic data and filtered educational content, an
approach motivated by the ``textbooks are all you need'' philosophy.
This regime prioritises structured reasoning and instruction
adherence; however, the model's strict chat template format
(\texttt{<|user|>...<|end|>}) may create sensitivity to prompts
that deviate from this expected structure, contributing to the
elevated UNKNOWN rate observed on MedQA.

Llama 3.2 \citep{grattafiori2024llama} (3B, Meta) is derived from
Llama 3.1 via pruning and knowledge distillation from larger Llama
3.1 models (8B and 70B), followed by multiple rounds of post-training
alignment using SFT, Rejection Sampling (RS), and DPO. This
multi-round alignment process, explicitly optimising dialogue
helpfulness across varied instruction formats, likely contributes
to the model's strong instruction-following performance and its
low UNKNOWN rates across all datasets.

Gemma 2 \citep{team2024gemma} (2B, Google DeepMind) is trained via
knowledge distillation from a substantially larger teacher model
rather than standard next-token prediction, using on-policy
distillation to mitigate train-inference distribution mismatch.
Post-training applies SFT, RLHF with a conversational reward model,
and model merging across hyperparameter configurations. Knowledge
distillation at this scale may instil stable but potentially
miscalibrated response tendencies inherited from the teacher model,
providing a mechanistic account of Gemma 2's high consistency scores
alongside its low accuracy: the model may be replicating confident
response patterns from its teacher that do not generalise correctly
to clinical reasoning tasks.

Mistral \citep{jiang2023mistral} (7B, Mistral AI) uses sliding
window attention and grouped-query attention for efficient inference,
and is instruction-tuned using SFT on publicly available conversation
datasets. Compared to Llama 3.2 and Phi-3, Mistral's instruct
fine-tuning used a more limited set of public instruction data and
requires strict \texttt{[INST]...[/INST]} delimiters. Prompts that
do not conform to this format may partially escape instruction
alignment, potentially contributing to higher UNKNOWN rates relative
to Llama 3.2 and Gemma 2.

Meditron-7B \citep{chen2023meditron} (EPFL) is built on Llama-2
and further pretrained on a curated medical corpus comprising PubMed
abstracts and international clinical practice guidelines. Critically,
Meditron-7B is a continued-pretraining model without instruction
tuning; it has not been trained to follow structured task instructions
or produce constrained single-token outputs. Its inclusion serves as
an architectural control condition: by contrasting a
domain-knowledgeable but non-instruction-tuned model against
instruction-tuned general models, this study isolates
instruction-following capability as a distinct and necessary
architectural requirement for structured clinical question answering,
independent of domain knowledge.

All models were selected because they are fully open-source, capable
of running on consumer CPU hardware without GPU acceleration, and
represent distinct model families with meaningfully different
training methodologies, enabling cross-architecture comparison of
consistency behaviour. All inference was conducted locally using
Ollama without domain-specific fine-tuning, ensuring that observed
consistency patterns reflect inherent model behaviour rather than
task-specific adaptation.

\begin{table}[H]
\centering
\small
\caption{Architectural and training characteristics of evaluated
models. SFT = Supervised Fine-Tuning; DPO = Direct Preference
Optimisation; RLHF = Reinforcement Learning from Human Feedback;
RS = Rejection Sampling; KD = Knowledge Distillation.
$\dagger$Meditron-7B is a continued pretraining model without
instruction tuning; its chat template is effectively undefined.}
\label{tab:model_arch}
\resizebox{\textwidth}{!}{%
\begin{tabular}{lcccccc}
\toprule
\textbf{Model} &
\textbf{Params} &
\textbf{Base Model} &
\textbf{Inst.\ Tuned} &
\textbf{Fine-Tuning Method} &
\textbf{Medical Domain} &
\textbf{Chat Template} \\
\midrule
Phi-3 Mini            & 3.8B & Phi-3 (Microsoft)    & Yes & SFT + DPO + RLHF                                  & No  & \texttt{<|user|>..<|end|>} \\
Llama 3.2             & 3B   & Llama 3.1 (Meta)     & Yes & SFT + RS + DPO (multi-round)                      & No  & \texttt{<|start\_header\_id|>..<|eot\_id|>} \\
Gemma 2               & 2B   & Gemma 2 (Google)     & Yes & KD + SFT + RLHF + Model Merging                   & No  & \texttt{<start\_of\_turn>..<end\_of\_turn>} \\
Mistral 7B            & 7B   & Mistral (Mistral AI) & Yes & SFT (public datasets)                             & No  & \texttt{[INST]..[/INST]} \\
Meditron-7B$\dagger$  & 7B   & Llama-2 (Meta)       & No  & Continued pretraining only (PubMed + guidelines)  & Yes & None \\
\bottomrule
\end{tabular}}
\end{table}

\subsection{Prompt Variation Design}

To systematically evaluate how small language models respond to 
differently worded clinical prompts, a prompt variation engine 
was designed to generate five semantically equivalent but 
stylistically distinct prompt formulations for each question.

\begin{itemize}
    \item \textbf{Original:} The question exactly as it appears in 
    the dataset, serving as the baseline condition.
    
    \item \textbf{Formal:} Rephrased in clinical academic language 
    to simulate how a healthcare professional might query a model 
    in a structured setting.
    
    \item \textbf{Simplified:} Written in plain everyday language, 
    simulating how a non-expert or patient might pose the same 
    question.
    
    \item \textbf{Roleplay:} The model is instructed to respond as 
    a practicing physician, testing whether persona-based prompting 
    affects answer consistency.
    
    \item \textbf{Direct:} Presents only the bare question and 
    answer options with minimal framing, assessing model behavior 
    without explicit instruction.
\end{itemize}

Each question across all three datasets was transformed into these 
five prompt styles, yielding 3,000 total prompts (600 questions 
$\times$ 5 styles). By keeping semantic content identical across 
variations, any differences in model output can be attributed to 
prompt sensitivity rather than differences in question content.

\subsection{Inference Setup}

All models were served locally via Ollama and queried through its 
REST API. Inference was conducted with temperature set to 0 and a 
maximum token limit of 10 tokens per response. Setting temperature 
to 0 ensures deterministic outputs, meaning that identical prompts 
produce identical responses across runs. This configuration isolates 
prompt variation as the primary source of output differences rather than 
stochastic sampling effects. Limiting output to 10 tokens prevents 
verbose responses and constrains the model to produce the single-letter 
(A, B, C, or D) or keyword (yes, no, or maybe) answer required by each 
dataset format.

Model responses were parsed using regular expression extraction. 
For multiple-choice datasets (MedQA and MedMCQA), the extractor 
identified the first standalone letter A, B, C, or D in the response. 
For PubMedQA, the extractor identified the first occurrence of 
\texttt{yes}, \texttt{no}, or \texttt{maybe}. Responses that did not 
contain a valid answer were categorized as \texttt{UNKNOWN}, 
representing instruction-following failures where the model does not produce a valid answer option. All responses, including 
\texttt{UNKNOWN} outputs, were retained for consistency scoring and 
subsequent analysis. All inference parameters were held constant across 
models and prompt styles to ensure fair comparison.

\subsection{Evaluation Metrics}

Four complementary metrics were defined to quantify model behavior 
under prompt variation.

\textbf{Consistency Score.} For each question, the majority answer 
was identified, defined as the response selected most frequently 
across the five prompt styles, and the proportion of prompt styles 
that agreed with it was computed. Formally, given responses 
$r_1, r_2, r_3, r_4, r_5$ for a question, the consistency score 
was defined as:

\begin{equation}
    C = \frac{\sum_{i=1}^{5} \mathbf{1}[r_i = \hat{r}]}{5}
\end{equation}

where $\hat{r}$ denotes the majority answer and $\mathbf{1}[\cdot]$ 
is the indicator function. A score of 1.0 indicates perfect 
consistency across all prompt styles, while a score of 0.2 indicates 
maximum inconsistency. For example, responses [A, B, B, B, A] 
yield a majority answer of B and a consistency score of $3/5 = 0.60$.

\textbf{Overall Accuracy.} The majority answer for each question 
is compared with the dataset ground-truth label, reflecting 
task-level correctness while accounting for variability across 
prompt styles.

\textbf{Per-Style Accuracy.} Accuracy is computed separately 
for each prompt style (original, formal, simplified, roleplay, 
direct), enabling identification of which prompt formulations 
the model handles most and least reliably.

\textbf{UNKNOWN Rate.} The proportion of responses that do not 
contain a valid answer option (A--D or yes/no/maybe), representing 
instruction-following failures that undermine clinical reliability.

Statistical significance was assessed using the Wilcoxon 
signed-rank test for comparisons of consistency scores and 
McNemar's test for paired accuracy comparisons. All p-values 
are reported without adjustment for multiple comparisons. 
Because the analysis is exploratory and hypothesis-generating, 
no correction for multiple testing was applied, consistent with 
prior exploratory clinical AI benchmarking studies 
\citep{bedi2025systematic}. Individual p-values should therefore 
be interpreted cautiously rather than as confirmatory statistical 
evidence.

Together, these metrics provide a holistic assessment of model 
behavior. A model may be highly consistent yet systematically 
incorrect, accurate only under specific prompt styles, or prone 
to instruction-following failures. Evaluating any single metric 
in isolation would obscure these clinically relevant failure modes.

\section{Results}\label{sec:results}

\subsection{Overall Consistency and Accuracy}

\begin{table}[H]
\centering
\small
\caption{Summary of consistency scores, overall accuracy, and 
instruction-following failure rates across all models and datasets. 
The best value among instruction-tuned models is shown in \textbf{bold}. 
$\dagger$Meditron-7B is a pretraining model without instruction 
tuning; high UNKNOWN rates reflect instruction-following failure 
rather than lack of domain knowledge.}
\label{tab:results}
\resizebox{\textwidth}{!}{
\begin{tabular}{llccccccccc}
\toprule
\textbf{Dataset} & \textbf{Model} & \textbf{Mean} & \textbf{Fully} & \textbf{Overall} & \textbf{UNKNOWN} & \textbf{Acc.} & \textbf{Acc.} & \textbf{Acc.} & \textbf{Acc.} & \textbf{Acc.} \\
 & & \textbf{Cons.} & \textbf{Cons.} & \textbf{Acc.} & \textbf{Rate} & \textbf{Orig.} & \textbf{Form.} & \textbf{Simp.} & \textbf{Role.} & \textbf{Dir.} \\
 & & & \textbf{(\%)} & \textbf{(\%)} & \textbf{(\%)} & \textbf{(\%)} & \textbf{(\%)} & \textbf{(\%)} & \textbf{(\%)} & \textbf{(\%)} \\
\midrule
\multirow{5}{*}{MedQA}
 & Phi-3 Mini  & 0.698 & 19.5 & 48.0 & 10.5 & 44.5 & 44.5 & 42.5 & 26.5 & 48.0 \\
 & Llama 3.2   & 0.776 & 35.0 & \textbf{49.0} & \textbf{0.8} & 47.5 & 44.5 & 46.0 & 47.0 & 43.5 \\
 & Gemma 2     & \textbf{0.888} & \textbf{63.5} & 40.0 & 2.1 & 39.0 & 39.0 & 37.0 & 35.0 & 35.0 \\
 & Mistral 7B  & 0.800 & 41.0 & 45.0 & 4.7 & 45.5 & 41.5 & 43.0 & 34.0 & \textbf{50.5} \\
 & Meditron$\dagger$ & 0.761 & 10.0 & 30.0 & 22.8 & 4.0 & 26.5 & 28.0 & 29.5 & 30.5 \\
\midrule
\multirow{5}{*}{MedMCQA}
 & Phi-3 Mini  & 0.730 & 27.0 & 53.0 & 4.6 & 52.0 & 49.5 & 48.5 & 37.5 & 52.0 \\
 & Llama 3.2   & 0.774 & 38.5 & \textbf{55.5} & \textbf{1.5} & \textbf{56.5} & 50.5 & 51.5 & 48.0 & 53.0 \\
 & Gemma 2     & \textbf{0.851} & \textbf{56.5} & 43.5 & 0.9 & 42.5 & 39.0 & 38.0 & 40.5 & 44.0 \\
 & Mistral 7B  & 0.812 & 45.5 & 45.5 & 4.7 & 44.5 & 43.0 & 42.5 & 41.5 & 44.0 \\
 & Meditron$\dagger$ & 0.762 & 8.0 & 33.5 & 22.4 & 4.0 & 33.0 & 30.0 & 33.5 & 29.5 \\
\midrule
\multirow{5}{*}{PubMedQA}
 & Phi-3 Mini  & 0.830 & 41.0 & 48.0 & 0.0 & 42.5 & \textbf{53.5} & 52.5 & 47.5 & 31.0 \\
 & Llama 3.2   & 0.807 & 45.5 & \textbf{65.0} & 2.8 & \textbf{59.5} & 51.0 & \textbf{64.5} & \textbf{65.0} & 56.5 \\
 & Gemma 2     & \textbf{0.845} & \textbf{50.5} & 33.0 & \textbf{1.3} & 28.0 & 41.5 & 47.5 & 27.0 & 30.5 \\
 & Mistral 7B  & 0.825 & 49.5 & 42.5 & 7.2 & 39.5 & 39.5 & 45.5 & 34.5 & 45.5 \\
 & Meditron$\dagger$ & 0.010 & 0.0 & 2.0 & 99.0 & 0.0 & 0.0 & 1.5 & 0.0 & 0.5 \\
\bottomrule
\end{tabular}
}

\end{table}

Statistical significance of observed differences was assessed 
using the Wilcoxon signed-rank test for consistency scores 
and McNemar's test for accuracy comparisons. Consistency 
differences between models were statistically significant 
on MedQA and MedMCQA (p$<$0.001 for most pairwise comparisons) 
but not on PubMedQA, suggesting that dataset characteristics 
influence the degree of prompt sensitivity. Accuracy 
differences were generally non-significant on MedQA, 
whereas on PubMedQA, Llama 3.2 significantly outperformed 
both Phi-3 Mini and Gemma 2 (p$<$0.001), indicating stronger 
performance on biomedical evidence-based questions. Full 
significance test results are provided in Table~\ref{tab:stats}.
\begin{longtable}{llllcc}
\caption{Pairwise statistical significance tests. Wilcoxon 
signed-rank test for consistency; McNemar test for accuracy.
*** p$<$0.001, ** p$<$0.01, * p$<$0.05, ns = not significant.
$\dagger$Meditron PubMedQA comparisons reflect instruction-following 
failure (99\% UNKNOWN rate).}
\label{tab:stats}\\
\toprule
\textbf{Dataset} & \textbf{M1} & \textbf{M2} & 
\textbf{Metric} & \textbf{p} & \textbf{Sig.} \\
\midrule
\endfirsthead
\toprule
\textbf{Dataset} & \textbf{M1} & \textbf{M2} & 
\textbf{Metric} & \textbf{p} & \textbf{Sig.} \\
\midrule
\endhead
\midrule
\multicolumn{6}{r}{\small\itshape Continued on next page} \\
\endfoot
\bottomrule
\endlastfoot
MedQA & Llama 3.2 & Phi-3 Mini & Cons. & 0.0005 & *** \\
 & Gemma 2 & Phi-3 Mini & Cons. & $<$.0001 & *** \\
 & Mistral & Phi-3 Mini & Cons. & $<$.0001 & *** \\
 & Llama 3.2 & Gemma 2 & Cons. & $<$.0001 & *** \\
 & Meditron & Phi-3 Mini & Cons. & $<$.0001 & *** \\
 & Meditron & Llama 3.2 & Cons. & 0.3342 & ns \\
 & Llama 3.2 & Phi-3 Mini & Acc. & 0.9020 & ns \\
 & Gemma 2 & Phi-3 Mini & Acc. & 0.0689 & ns \\
 & Mistral & Phi-3 Mini & Acc. & 0.5383 & ns \\
 & Llama 3.2 & Gemma 2 & Acc. & 0.0512 & ns \\
 & Meditron & Phi-3 Mini & Acc. & 0.0004 & *** \\
 & Meditron & Llama 3.2 & Acc. & 0.0001 & *** \\
\midrule
MedMCQA & Llama 3.2 & Phi-3 Mini & Cons. & 0.0437 & * \\
 & Gemma 2 & Phi-3 Mini & Cons. & $<$.0001 & *** \\
 & Mistral & Phi-3 Mini & Cons. & 0.0001 & *** \\
 & Llama 3.2 & Gemma 2 & Cons. & 0.0001 & *** \\
 & Meditron & Phi-3 Mini & Cons. & 0.0007 & *** \\
 & Meditron & Llama 3.2 & Cons. & 0.2052 & ns \\
 & Llama 3.2 & Phi-3 Mini & Acc. & 0.5962 & ns \\
 & Gemma 2 & Phi-3 Mini & Acc. & 0.0327 & * \\
 & Mistral & Phi-3 Mini & Acc. & 0.1060 & ns \\
 & Llama 3.2 & Gemma 2 & Acc. & 0.0075 & ** \\
 & Meditron & Phi-3 Mini & Acc. & 0.0002 & *** \\
 & Meditron & Llama 3.2 & Acc. & $<$.0001 & *** \\
\midrule
PubMedQA & Llama 3.2 & Phi-3 Mini & Cons. & 0.1055 & ns \\
 & Gemma 2 & Phi-3 Mini & Cons. & 0.8497 & ns \\
 & Mistral & Phi-3 Mini & Cons. & 0.6172 & ns \\
 & Llama 3.2 & Gemma 2 & Cons. & 0.1055 & ns \\
 & Meditron$\dagger$ & Phi-3 Mini & Cons. & $<$.0001 & *** \\
 & Meditron$\dagger$ & Llama 3.2 & Cons. & $<$.0001 & *** \\
 & Llama 3.2 & Phi-3 Mini & Acc. & $<$.0001 & *** \\
 & Gemma 2 & Phi-3 Mini & Acc. & 0.0002 & *** \\
 & Mistral & Phi-3 Mini & Acc. & 0.1930 & ns \\
 & Llama 3.2 & Gemma 2 & Acc. & $<$.0001 & *** \\
 & Meditron & Phi-3 Mini & Acc. & $<$.0001 & *** \\
 & Meditron & Llama 3.2 & Acc. & $<$.0001 & *** \\
\end{longtable}

\begin{figure}[H]
    \centering
    \includegraphics[width=0.85\textwidth]{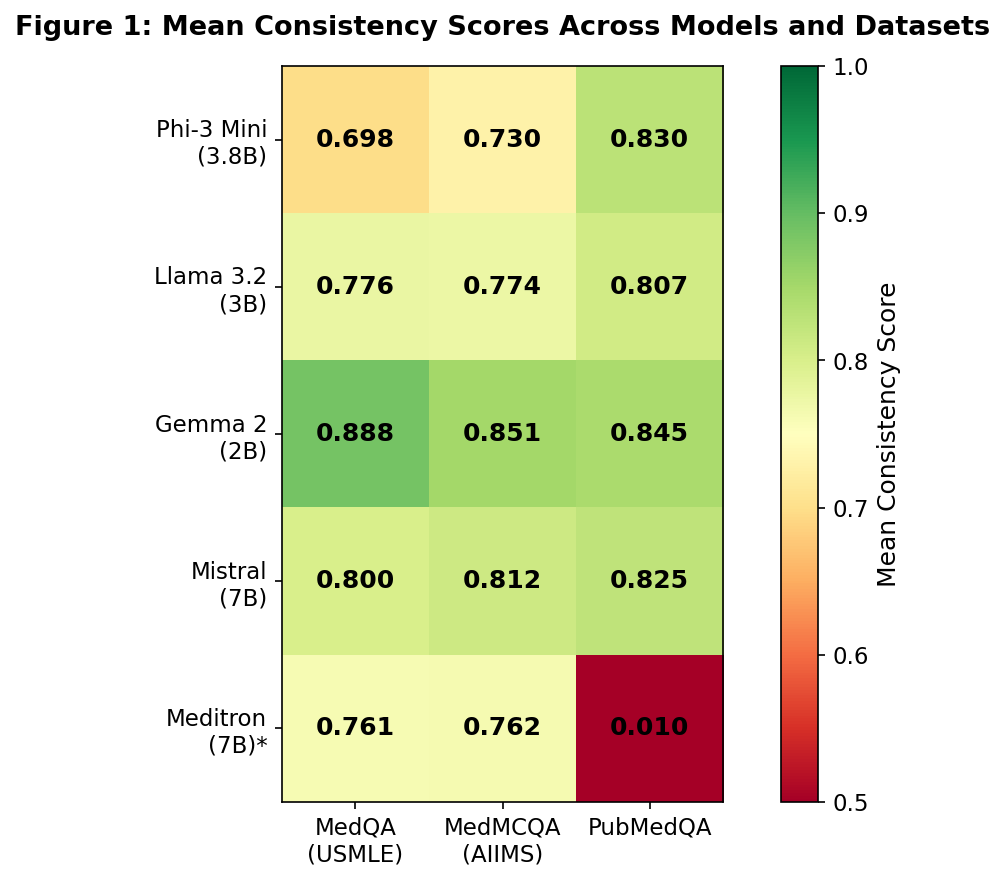}
    \caption{Mean consistency scores across models and datasets. 
    Higher scores indicate greater agreement across prompt styles.}
    \label{fig:consistency}
\end{figure}

Across all three datasets, consistency scores varied across models 
without a clear correlation with model size (Figure~\ref{fig:consistency}). 
Phi-3 Mini (3.8B) exhibited the lowest mean consistency scores 
(0.698--0.830), while Gemma 2 (2B) achieved the highest 
(0.845--0.888) despite being the smallest model tested. Mistral 7B 
demonstrated strong but slightly variable consistency (0.800--0.825), 
and Llama 3.2 showed intermediate consistency (0.774--0.807).

\begin{figure}[H]
    \centering
    \includegraphics[width=0.85\textwidth]{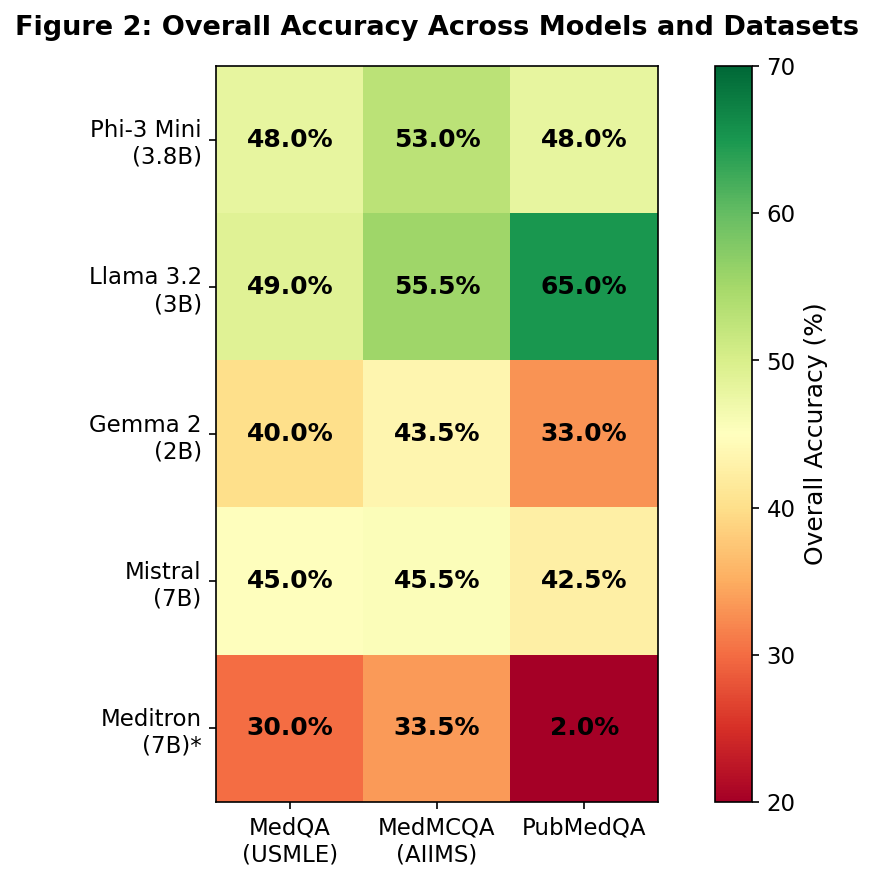}
    \caption{Overall accuracy across models and datasets, calculated 
    using majority answer against ground truth labels.}
    \label{fig:accuracy}
\end{figure}

Accuracy patterns diverged substantially from consistency rankings 
(Figure~\ref{fig:accuracy}). Llama 3.2 achieved the highest overall 
accuracy across datasets (49.0--65.0\%), followed by Phi-3 Mini 
(48.0--53.0\%) and Mistral 7B (42.5--45.5\%). Gemma 2, despite 
exhibiting the highest consistency scores, achieved the lowest 
accuracy (33.0--43.5\%). This inverse relationship between 
consistency and accuracy in Gemma 2 highlights a critical finding: 
high consistency does not imply high accuracy. Models can be 
\textit{reliably wrong}, producing the same incorrect answer 
across all prompt variations, which represents a particularly 
dangerous failure mode in clinical decision support contexts.

Meditron-7B, included as a domain-adapted pretraining model 
without instruction tuning, exhibited markedly different 
behavior from all instruction-tuned models. While it 
achieved moderate consistency on MedQA and MedMCQA 
(0.761 and 0.762, respectively), its UNKNOWN rates were 
substantially higher (22.8\% and 22.4\%), indicating 
frequent instruction-following failures. On PubMedQA, 
Meditron-7B produced near-complete failure, with a 
consistency score of 0.010, 2.0\% accuracy, and a 99.0\% 
UNKNOWN rate. This suggests that the model is unable to 
produce structured yes/no/maybe responses without instruction 
tuning. These results indicate that domain-specific pretraining 
alone is insufficient for structured clinical question answering 
and that instruction-following capability is a prerequisite for 
clinical deployment.

Due to the extremely high UNKNOWN rate observed for Meditron-7B 
on PubMedQA (99.0\%), statistical comparisons involving this model 
on that dataset should be interpreted cautiously, as the 
near-absence of valid responses renders consistency metrics 
unreliable.

\subsection{Effect of Prompt Style on Accuracy}

\begin{figure}[H]
    \centering
    \includegraphics[width=\textwidth]{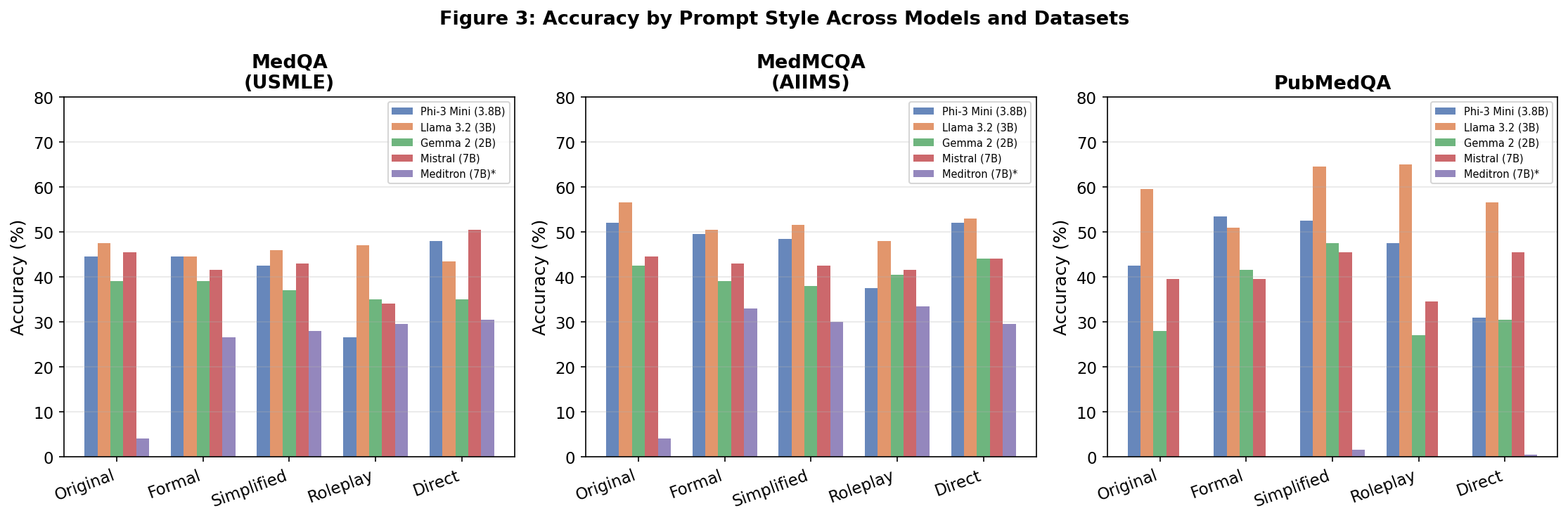}
    \caption{Accuracy by prompt style across models and datasets. 
    Roleplay consistently underperforms relative to other styles.}
    \label{fig:style_accuracy}
\end{figure}

\begin{figure}[H]
    \centering
    \includegraphics[width=\textwidth]{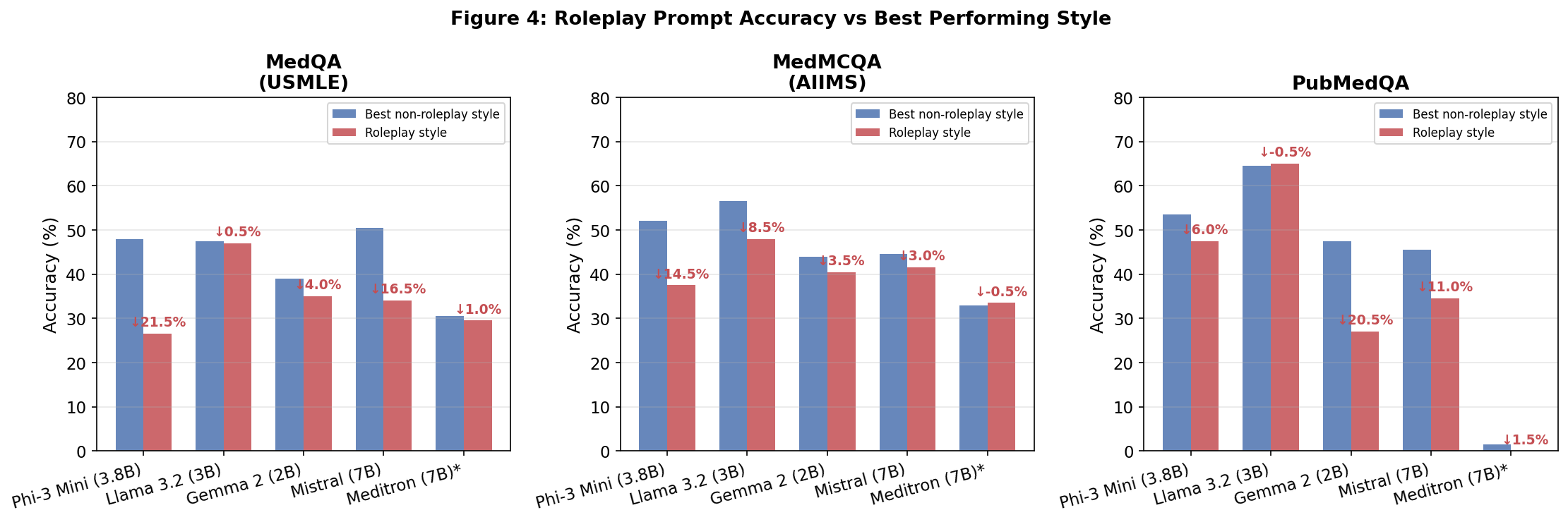}
    \caption{Roleplay prompt accuracy versus best performing 
    non-roleplay style across all models and datasets, with 
    annotated performance gaps.}
    \label{fig:roleplay_gap}
\end{figure}

Prompt style demonstrated a consistent and measurable effect on 
model accuracy across all five models and three datasets 
(Figure~\ref{fig:style_accuracy}). The Roleplay prompt style, 
in which models were instructed to respond as a practicing 
physician, consistently underperformed relative to all other 
styles. The most pronounced decline was observed in Phi-3 Mini, 
which achieved 48.0\% accuracy under the Direct style but only 
26.5\% under the Roleplay style on MedQA, representing a drop 
of 21.5 percentage points. Similar reductions were observed 
across Llama 3.2, Gemma 2, and Mistral 7B, suggesting that 
roleplay prompting is systematically detrimental rather than 
model-specific (Figure~\ref{fig:roleplay_gap}).

In contrast, the Direct and Original prompt styles produced 
the highest and most stable accuracy across models and datasets, 
suggesting that minimal framing is preferable for reliable 
clinical question answering. These findings carry direct 
implications for healthcare AI deployment: prompt formulations 
intended to simulate clinical expertise, such as persona-based 
instructions, may paradoxically reduce the reliability of 
small language model outputs in patient-facing applications.

\subsection{Instruction-Following Failure Rate}

\begin{figure}[H]
    \centering
    \includegraphics[width=0.85\textwidth]{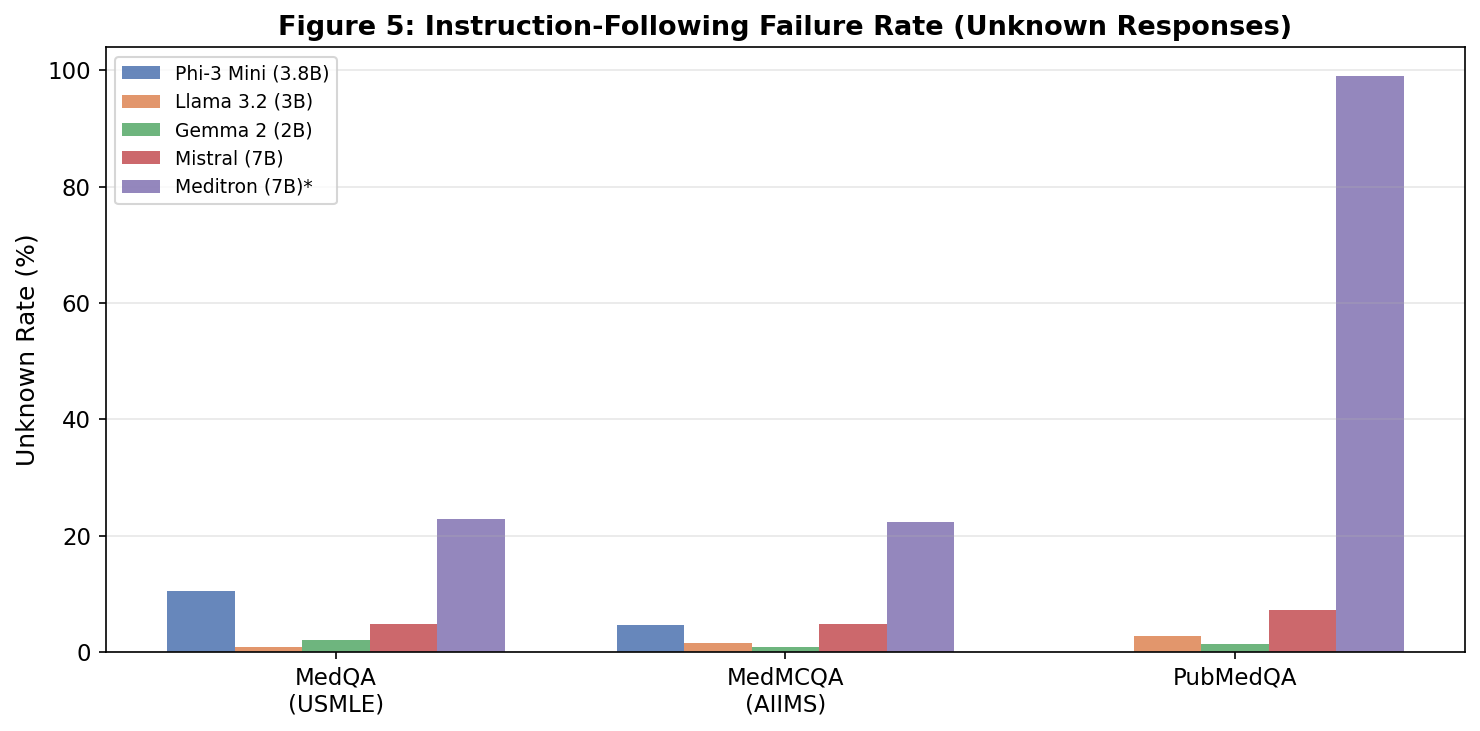}
    \caption{Instruction-following failure rates (UNKNOWN responses) 
    across models and datasets. Phi-3 Mini shows the highest failure 
    rate on MedQA at 10.5\%.}
    \label{fig:unknown}
\end{figure}

The UNKNOWN rate, defined as the proportion of responses that did not 
contain a valid answer option, varied substantially across models 
and datasets (Figure~\ref{fig:unknown}). Phi-3 Mini exhibited the 
highest UNKNOWN rate on MedQA at 10.5\%, indicating that 
approximately one in ten queries failed to produce a usable 
response. Mistral 7B, despite being the largest model evaluated, 
also demonstrated non-negligible UNKNOWN rates across all three 
datasets (4.7\%--7.2\%), suggesting that model size alone does not 
guarantee reliable instruction following.

In contrast, Llama 3.2 and Gemma 2 achieved the lowest UNKNOWN 
rates across datasets (0.8\%--2.8\% and 0.9\%--2.1\%, 
respectively), consistently producing responses in the expected 
format. These results suggest that instruction adherence is a 
model-specific characteristic independent of parameter count.

Meditron-7B exhibited the most severe instruction-following 
failures of all models evaluated, with UNKNOWN rates of 
22.8\% on MedQA, 22.4\% on MedMCQA, and 99.0\% on 
PubMedQA. The near-complete failure on PubMedQA, where 
99 of 100 responses were invalid, is consistent with 
Meditron's architecture as a base pretraining model 
without instruction tuning. Unlike the instruction-following 
failures observed in Phi-3 Mini, which appear to be 
prompt-style dependent, Meditron's failures reflect a 
fundamental absence of structured response capability 
rather than prompt sensitivity.

From a clinical deployment perspective, high UNKNOWN rates 
represent a critical reliability concern. A model that 
frequently fails to produce a valid response cannot serve 
as a dependable decision support tool, as such failures 
interrupt clinical workflows, introduce uncertainty, and 
erode clinician trust in AI-assisted systems.

The variation in UNKNOWN rates across instruction-tuned models
also reflects differences in the scope and methodology of their
post-training alignment. Llama 3.2 and Gemma 2, which produced
the lowest UNKNOWN rates, underwent multi-round alignment processes
optimising across diverse instruction formats. Mistral 7B and
Phi-3 Mini, which showed higher UNKNOWN rates on certain datasets,
relied on more constrained instruction fine-tuning that may be
more sensitive to prompt phrasings that diverge from their expected
template structures. These patterns suggest that the breadth of
instruction format coverage during fine-tuning, not merely
parameter count, is the primary architectural determinant of
instruction-following reliability in structured clinical tasks.

\subsection{Consistency Versus Accuracy Relationship}

\begin{figure}[H]
    \centering
    \includegraphics[width=\textwidth]{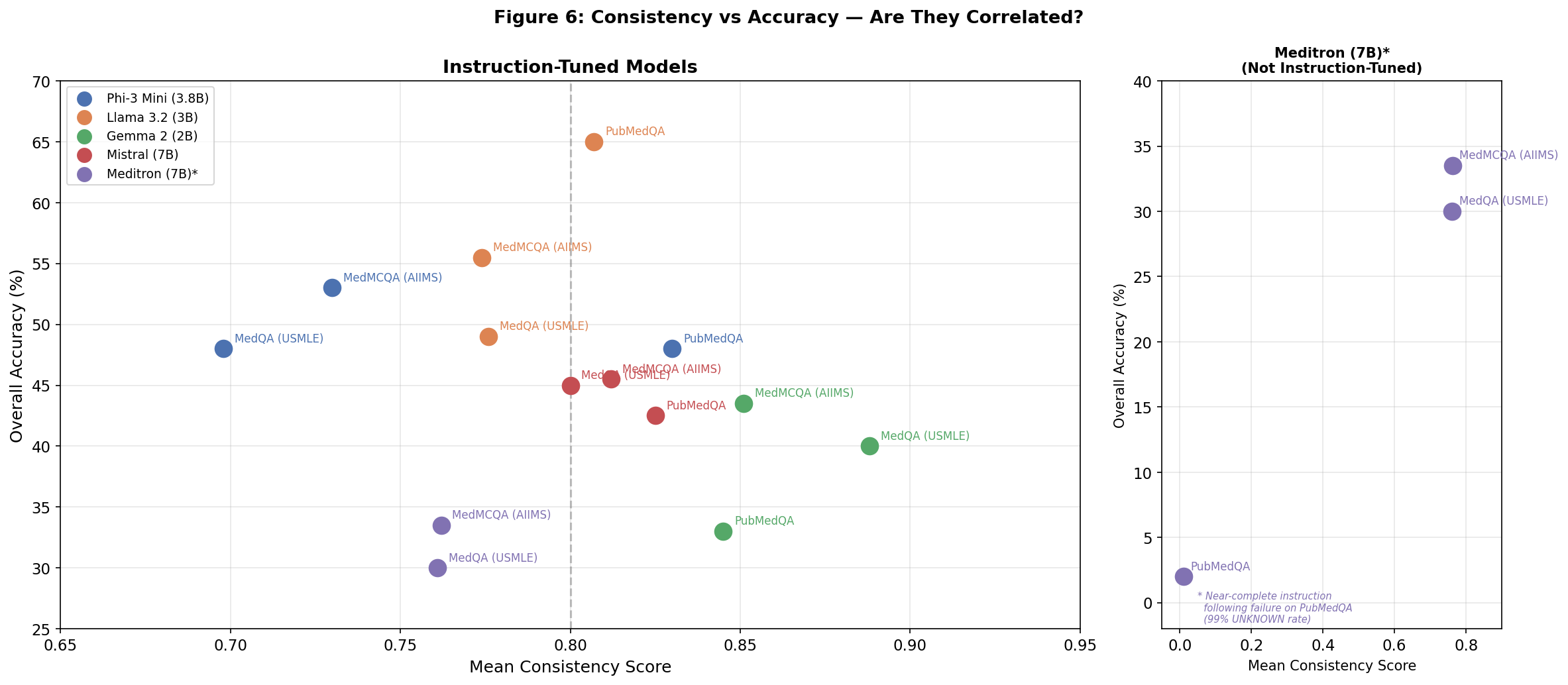}
    \caption{Scatter plot of mean consistency score versus overall 
    accuracy across all models and datasets. No clear positive 
    correlation is observed, indicating that consistency and 
    accuracy are independent metrics.}
    \label{fig:scatter}
\end{figure}

Analysis of the relationship between mean consistency scores and 
overall accuracy revealed no clear positive correlation across 
models and datasets (Figure~\ref{fig:scatter}). Gemma 2 
exemplifies this disconnect most clearly: it achieved the highest 
consistency scores across all datasets (0.845--0.888) while 
simultaneously recording the lowest accuracy (33.0--43.5\%), 
indicating that it consistently produces incorrect answers with 
high confidence. Conversely, Llama 3.2 demonstrated moderate 
consistency (0.774--0.807) yet achieved the highest accuracy 
across most datasets (49.0--65.0\%), showing that a model can 
be more frequently correct even when its answers vary across 
prompt styles.

\begin{figure}[H]
    \centering
    \includegraphics[width=\textwidth]{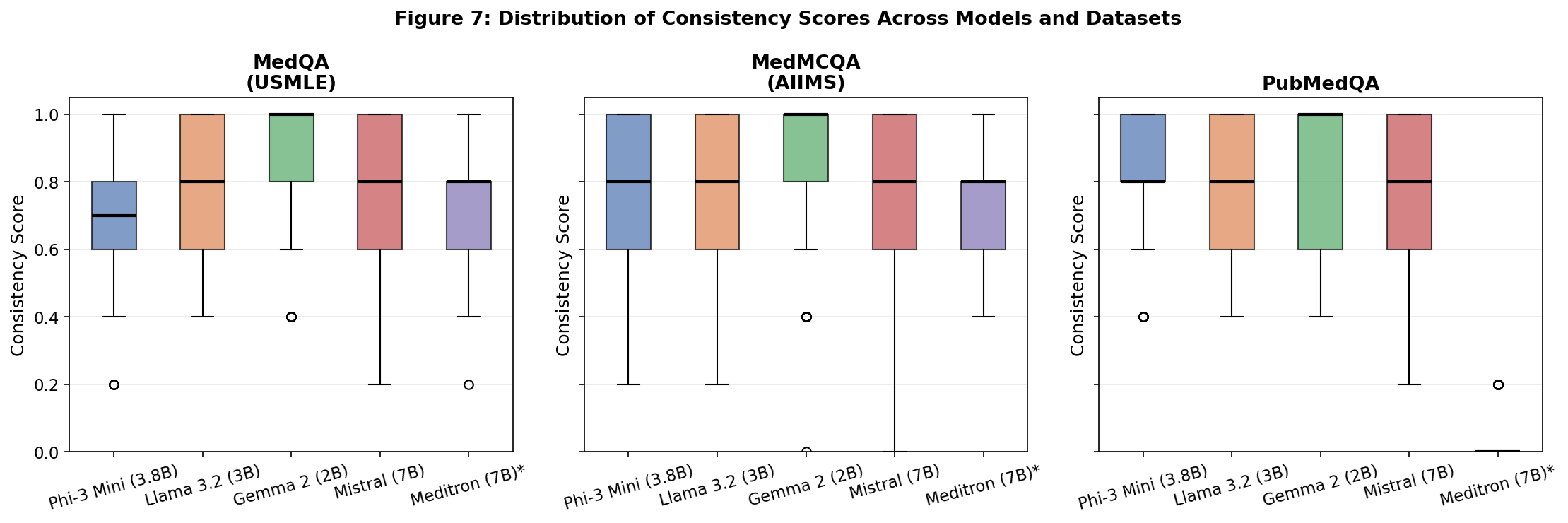}
    \caption{Distribution of consistency scores across models and 
    datasets. Phi-3 Mini shows the widest spread on MedQA; 
    Gemma 2 shows a tight but high distribution reflecting 
    systematic consistency without accuracy.}
    \label{fig:boxplot}
\end{figure}

The distribution of consistency scores further illustrates 
model-level differences in reliability 
(Figure~\ref{fig:boxplot}). Phi-3 Mini exhibited the widest 
spread of consistency scores on MedQA, indicating substantial 
variability in prompt sensitivity across questions. Gemma 2 
showed a tighter, higher distribution, remaining consistently 
stable but systematically inaccurate.

These findings demonstrate that consistency and accuracy are 
independent dimensions of model performance. Evaluating either 
metric in isolation is insufficient for clinical AI assessment. 
A model that is highly consistent but consistently wrong may 
be more dangerous than one that occasionally contradicts itself, 
as the former provides false confidence in systematically 
incorrect outputs. Safe clinical deployment therefore requires 
joint evaluation of both consistency and accuracy.

\section{Discussion}\label{sec:discussion}

From a clinical informatics perspective, these findings highlight the importance of 
evaluating AI systems not only for accuracy but also for response stability. Clinical 
decision support tools are often queried in multiple ways by clinicians, and inconsistent 
responses to semantically equivalent queries may undermine trust in AI-assisted systems. 
Conversely, consistently incorrect outputs may create a false sense of reliability if 
repeated answers are interpreted as evidence of model confidence. Evaluation frameworks 
that jointly assess accuracy and response stability therefore provide a more comprehensive 
basis for assessing whether language models are sufficiently reliable for integration into 
clinical workflows.

\subsection{Consistency and Accuracy are Independent Metrics}

A central empirical finding of this study is that consistency and accuracy 
represent independent dimensions of small language model performance 
in clinical question answering. While prior evaluations of clinical 
AI have predominantly focused on accuracy as the primary metric 
\citep{bedi2025systematic}, the results presented here demonstrate 
that a model can exhibit high consistency while still producing 
systematically incorrect answers, a failure mode referred to in 
this work as \textit{reliable incorrectness}.

Gemma 2 illustrates this most clearly, achieving the highest
mean consistency scores across all datasets (0.845--0.888) while
recording the lowest overall accuracy (33.0--43.5\%). This pattern
suggests that the model has learned stable response tendencies that
are nonetheless clinically incorrect. Across 200 MedQA questions,
Gemma 2 produced perfectly consistent yet incorrect answers on 77
questions (38.5\%), selecting the same wrong option across all five
prompt styles with a consistency score of 1.0.

The danger of reliable incorrectness is both psychological and 
technical. Consider a primary care physician using an AI assistant 
to support diagnostic reasoning. If the model returns the same 
answer regardless of how a question is phrased, this repetition 
creates an illusion of reliability. A clinician might reason, 
``it keeps suggesting pulmonary embolism; it must be confident.'' 
That apparent stability can feel reassuring. However, if the correct 
diagnosis is pneumonia, the model's consistency actively reinforces 
a false conclusion. Over time, clinicians may develop misplaced 
trust in systems that appear stable and decisive without recognising 
that the stability reflects systematic error rather than clinical 
knowledge. In practice, this could translate into repeated 
misdiagnoses, inappropriate investigations, or delayed treatment, 
not because the model behaves randomly, but because it is 
confidently and predictably incorrect.

This finding aligns with Kim et al. \citep{kim2025hallucination}, 
who demonstrate that foundation models frequently produce 
hallucinations with high apparent confidence and that clinicians 
consider such outputs particularly dangerous because they are 
difficult to detect. The present results extend this concern to 
the domain of prompt sensitivity. A model that produces consistent 
but incorrect answers across varied prompt formulations compounds 
this risk by appearing robust to rephrasing.

These observations underscore that accuracy and consistency must 
be evaluated jointly in any clinical AI assessment framework. 
Neither metric alone is sufficient to characterize the safety 
profile of a model intended for healthcare deployment.

The architectural characteristics of Gemma 2 offer a plausible
mechanistic account of this pattern. The 2B Gemma 2 model is
trained via knowledge distillation from a substantially larger
teacher model, inheriting the teacher's response tendencies rather
than learning independently from ground-truth labels. If the
teacher model has developed systematic biases in clinical question
answering --- for instance, preferring certain option positions or
applying heuristics that generalise poorly to medical reasoning ---
these biases may be faithfully replicated in the distilled student
model, producing high cross-prompt consistency that reflects the
teacher's error patterns rather than genuine clinical knowledge.
This distillation-induced miscalibration represents a distinct
architectural risk factor for clinical AI deployment that
accuracy-only evaluation frameworks would not detect.

\subsection{Roleplay Prompts Consistently Underperform}

Across all five models and three datasets, roleplay-style prompts 
consistently produced lower accuracy than direct or original 
question formats. The most pronounced decline was observed in 
Phi-3 Mini on MedQA, where roleplay accuracy dropped 21.5 
percentage points below the best-performing style. Critically, 
this pattern was consistent across all model families tested, 
suggesting a systematic rather than model-specific phenomenon.

Roleplay prompts appear to introduce a form of task interference 
that is particularly detrimental for small language models. When 
a prompt begins with persona framing such as ``You are a senior 
physician taking a licensing examination,'' the model must 
simultaneously interpret and simulate a clinical identity while 
reasoning through a structured medical question.

For smaller models with limited representational capacity, 
this dual requirement may dilute attention from the core 
reasoning task. Furthermore, persona-based prompts may activate 
patterns learned from conversational or narrative training data 
rather than the structured exam-style reasoning required for 
clinical QA benchmarks, effectively shifting the task from 
question answering to performance-oriented text generation.

This finding has direct practical implications. Prompt engineering 
guidelines for clinical AI systems often recommend persona framing 
to make outputs sound more authoritative or clinically appropriate 
\citep{white2023prompt}. These results suggest that, for small 
open-source models, such framing may paradoxically reduce factual 
reliability. Developers building clinical decision support tools 
on small language models should favor minimal, direct prompt 
formulations over persona-based instructions.

This observation also extends findings from Ngweta et al. 
\citep{ngweta2024robustness}, who demonstrated that prompt 
format changes cause measurable performance variation in general 
NLP tasks. The present results show that, in clinical settings, 
one specific prompt format, roleplay, is systematically and 
substantially more harmful than others, representing a concrete 
and actionable finding for healthcare AI deployment.

This phenomenon is likely amplified by the prompt template
conventions enforced during instruction fine-tuning. Each model
evaluated uses a distinct chat template: Phi-3 Mini employs
\texttt{<|user|>...<|end|>} delimiters, Llama 3.2 uses
\texttt{<|start\_header\_id|>...<|eot\_id|>} markers, and
Mistral 7B requires \texttt{[INST]...[/INST]} framing. Roleplay
prompts, which begin with persona instructions such as ``You are
a senior physician,'' may partially misalign with the expected
input structure of these templates, shifting the model's processing
toward narrative or conversational generation rather than
constrained structured reasoning. This suggests that roleplay
sensitivity reflects not merely task framing but an interaction
between prompt style and the learned template expectations
embedded during instruction tuning --- an architectural rather
than purely behavioural phenomenon.

\subsection{Implications for Low-Resource Healthcare Deployment}

A primary motivation for this study was the practical challenge 
of deploying AI in resource-constrained healthcare environments. 
Many rural clinics, community hospitals, and health systems in 
low-income settings lack access to GPU infrastructure, reliable 
high-speed internet, or cloud-based AI services. In these 
contexts, locally deployable open-source models in the 2B--7B 
parameter range represent a more realistic and equitable 
alternative to large proprietary systems. The experimental design 
intentionally reflected these constraints: all models were 
evaluated on consumer CPU hardware without domain-specific 
fine-tuning, simulating realistic deployment conditions rather 
than idealized research environments.

These findings provide concrete guidance for model selection in 
such settings. Although Gemma 2 demonstrated the highest 
consistency scores, its substantially lower accuracy suggests 
that it may systematically mislead clinicians, particularly in 
low-oversight environments where AI outputs may not be routinely 
verified. Mistral 7B, despite being the largest model evaluated, 
did not clearly outperform smaller alternatives and exhibited 
unexpected instruction-following failures, suggesting that 
larger parameter counts do not automatically justify greater 
hardware demands.

Llama 3.2 (3B) emerged as the most balanced candidate under the 
evaluation criteria used in this study, achieving the highest 
overall accuracy (49.0--65.0\%), moderate consistency, and 
among the lowest instruction-following failure rates. For 
low-resource healthcare deployment scenarios where clinician 
oversight may be limited, a model offering strong accuracy and 
reliable instruction adherence is preferable to one that 
maximizes consistency at the expense of correctness.

More broadly, these results suggest that deployment decisions 
for clinical AI systems should be guided by joint evaluation of 
accuracy, consistency, and instruction compliance rather than 
any single metric. In settings where errors carry direct patient 
safety implications, the cost of systematic incorrectness, even 
when delivered consistently, outweighs the benefit of apparent 
stability. This study provides a practical evaluation framework 
for low-resource healthcare AI deployment that can be reproduced 
without specialized infrastructure, making it accessible to 
researchers and practitioners in the settings it is intended 
to serve.

\subsection{Domain Knowledge Versus Instruction Following}

The inclusion of Meditron-7B reveals an important distinction 
between two capabilities required for clinical AI deployment: 
domain knowledge and instruction adherence. Meditron-7B was 
pretrained on a curated medical corpus comprising PubMed 
articles and clinical guidelines \citep{chen2023meditron}, 
encoding substantial medical domain knowledge. However, 
without instruction tuning, it failed to produce structured 
responses in the required format across all three datasets, 
with UNKNOWN rates reaching 99.0\% on PubMedQA.

This finding suggests that domain knowledge and task usability 
are orthogonal capabilities. A model can encode rich medical 
knowledge through continued pretraining yet remain clinically 
unusable for structured question answering if it lacks the 
ability to follow task-specific instructions. For practitioners 
deploying AI in clinical settings, this distinction has direct 
practical implications. Selecting a model based solely on its 
training data, without verifying instruction-following 
capability, may result in a system that is knowledgeable but 
functionally unusable in structured workflows.

These observations also motivate a clear direction for future 
work: instruction-tuning Meditron or similar domain-pretrained 
models on structured clinical QA datasets and evaluating 
whether the combination of domain knowledge and instruction 
following produces models that are both more accurate and 
more consistent than general instruction-tuned alternatives.

From an architectural standpoint, this finding reflects a
fundamental distinction between two training objectives that are
often conflated in clinical AI discourse. Continued pretraining on
medical corpora, as applied to Meditron-7B, optimises the model's
internal representations of medical knowledge but does not instil
the structured output behaviour required for task-constrained
deployment. Instruction tuning, by contrast, explicitly trains the
model to produce constrained responses in specific formats in
response to specific prompt structures. These are orthogonal
capabilities: a model can encode rich biomedical knowledge through
pretraining yet remain clinically unusable because it lacks the
architectural conditioning to translate that knowledge into a
structured, deployable output. For practitioners selecting models
for low-resource clinical deployment, this distinction implies
that medical domain pretraining alone is an insufficient selection
criterion; instruction-following capability must be verified
independently.

\subsection{Limitations}

Several limitations should be considered when interpreting these
findings. These limitations define the scope of the current study
and collectively motivate the future directions described in
Section~\ref{sec:future}.

First, 200 questions per dataset were sampled, which provides a
computationally feasible and reproducible evaluation but may not
fully capture the diversity of each benchmark. Second, the three datasets used, 
MedQA, MedMCQA, and PubMedQA, focus primarily on structured 
examination-style and research-based questions and may not reflect the 
complexity of real-world clinical dialogue or unstructured patient 
interactions.

Third, human clinical evaluation was not conducted to assess whether 
consistent but incorrect model outputs would meaningfully influence 
physician decision-making in practice. Such evaluation would strengthen 
the clinical validity of these findings but was beyond the scope of this 
study. Fourth, all models were evaluated without domain-specific 
fine-tuning, which may underestimate the performance achievable through 
medical adaptation techniques such as instruction tuning or 
retrieval-augmented generation \citep{lewis2020rag}.

Fifth, the evaluation was restricted to multiple-choice and 
yes/no/maybe answer formats, whereas real clinical environments 
frequently involve open-ended reasoning, differential diagnosis, 
and contextual patient data. Sixth, hardware constraints limited 
the evaluation to models in the 2B--7B parameter range running on 
consumer CPU hardware, precluding direct comparison with larger 
proprietary systems such as GPT-4 or Claude. Seventh, while the 
five prompt styles were designed to simulate realistic variation 
in clinical query formulation, they do not exhaust the full range 
of phrasings a clinician might use in practice. Finally, 
Meditron-7B was evaluated in its base pretraining form without 
instruction tuning, which limits direct comparison with 
instruction-tuned models and may underrepresent the potential of 
domain-adapted models when properly instruction-tuned for 
structured clinical tasks.

Together, these limitations define the scope of this study as an
initial systematic analysis of prompt sensitivity under constrained
deployment conditions. The evaluation framework introduced here is
designed to be reproducible and extensible, supporting future
clinical validation studies without requiring specialised
infrastructure.

\subsection{Future Work}\label{sec:future}

Several directions emerge naturally from this study. First, 
future research should evaluate whether domain-specific 
fine-tuning of small models on medical corpora reduces prompt 
sensitivity and improves consistency. If fine-tuned models 
demonstrate stronger stability under prompt variation, this 
would suggest that training data alignment, rather than model 
architecture, is the primary driver of inconsistency.

In particular, instruction-tuning Meditron-7B or similar 
domain-pretrained models on structured clinical QA datasets 
represents a direct next step. This would test whether combining 
domain knowledge with instruction-following capability produces 
models that are both more accurate and more consistent than the 
general instruction-tuned alternatives evaluated in this study.

Second, integrating retrieval-augmented generation (RAG) 
pipelines with small clinical models represents a promising 
direction. RAG-augmented systems ground responses in retrieved 
evidence, which may reduce both factual hallucinations and 
prompt-induced answer variation. Evaluating whether external 
knowledge retrieval mitigates the consistency–accuracy tradeoff 
observed in this study would be a natural extension 
\citep{lewis2020rag}.

Third, evaluating larger open-source and proprietary models, 
when hardware resources permit, would clarify whether the 
independence between consistency and accuracy observed here 
persists at greater parameter scales or whether it is specific 
to sub-7B models.

Beyond technical extensions, human-centered validation is 
essential. A controlled clinical study involving practicing 
physicians could assess how consistent but incorrect model 
outputs influence clinical reasoning, trust calibration, and 
decision-making in realistic workflows. Such a study would 
strengthen the ecological validity of the consistency metric 
introduced here.

Translationally, these findings support the development of a 
lightweight clinical decision support prototype for general 
practitioners in low-resource settings, built around the model 
demonstrating the strongest balance of accuracy and instruction 
adherence, with prompt engineering guardrails informed by the 
sensitivity analysis. Extending this evaluation framework to 
multilingual datasets and non-English clinical benchmarks is 
also critical, as many low-resource healthcare environments 
operate outside English-dominant settings. Together, these 
directions would advance this work from controlled benchmarking 
toward safe, deployable clinical AI systems grounded in both 
technical rigor and real-world constraints.

\section{Conclusion}\label{sec:conclusion}

This study systematically evaluates prompt sensitivity in five open-source language models (2B--7B parameters), including four general instruction-tuned models and one domain-adapted pretraining model, across three clinical question answering datasets using five controlled prompt style variations, yielding 3,000 total evaluation instances.

Four principal findings emerge from this analysis. First, consistency and accuracy are independent metrics: models that achieve high consistency are not necessarily more accurate, with Gemma 2 demonstrating the most pronounced case of reliable incorrectness, achieving the highest consistency scores while recording the lowest accuracy across all datasets. Second, roleplay-style prompts consistently reduce accuracy across all models and datasets, suggesting that persona-based framing introduces systematic instability in clinical reasoning tasks and should be avoided in healthcare AI prompt design. Third, instruction-following failures vary substantially across models and are not determined by parameter count alone, indicating that model size is an insufficient proxy for deployment reliability under hardware constraints. Fourth, Meditron-7B, a domain-adapted model without instruction tuning, exhibited near-complete failure on structured clinical QA tasks despite encoding substantial medical domain knowledge, demonstrating that domain knowledge and instruction-following capability are orthogonal requirements for clinical AI deployment.

Taken together, these results demonstrate that consistency alone cannot serve as a proxy for clinical reliability. Models that appear stable under prompt variation may still pose significant patient safety risks if their answers are systematically incorrect. Safe deployment of AI in clinical settings, particularly in low-resource environments where human oversight may be limited, requires joint evaluation of accuracy, consistency, and instruction adherence.

More broadly, this work highlights the need for the research community to move beyond accuracy-only evaluation frameworks for clinical AI and adopt multidimensional assessment approaches that reflect the reliability requirements of real-world healthcare deployment. The evaluation framework, datasets, and code introduced in this study are publicly available at \citep{github2026} to support reproducibility and future research in this direction.

\section*{Acknowledgments}

This study received no funding. All experiments were conducted
on personal hardware by an independent researcher. The author
thanks the open-source communities behind Ollama, HuggingFace
Datasets, and the model developers whose publicly released
technical reports informed the architectural analysis in this
work.

\section*{Conflicts of Interest}

The author declares no conflicts of interest. No funding was
received for this study.

\section*{Data Availability}

The evaluation framework, scored datasets, and all code used
in this study are publicly available at
\url{https://github.com/shravani-01/clinical-llm-eval}
\citep{github2026}. The three clinical benchmarks used
(MedQA, MedMCQA, PubMedQA) are publicly available via
HuggingFace Datasets.

\section*{Ethics Statement}

This study used publicly available, de-identified benchmark
datasets (MedQA, MedMCQA, PubMedQA) and did not involve
human subjects, patient data, or any form of primary data
collection. No ethics board review was required.

\bibliographystyle{plainnat}
\bibliography{references}

@article{jin2021disease,
  title={{What Disease Does This Patient Have? {A} Large-Scale Open Domain
  Question Answering Dataset from Medical Exams}},
  author={Jin, Di and Pan, Eileen and Oufattole, Nassim and 
  Weng, Wei-Hung and Fang, Hanyi and Szolovits, Peter},
  journal={Applied Sciences},
  year={2021}
}

@inproceedings{pal2022medmcqa,
  title={{{MedMCQA}: {A} Large-Scale Multi-Subject Multi-Choice Dataset 
  for Medical Exam Comprehension}},
  author={Pal, Ankit and Umapathi, Logesh Kumar and Sankarasubbu, Malaikannan},
  booktitle={Conference on Health, Inference, and Learning},
  year={2022}
}

@inproceedings{jin2019pubmedqa,
  title={{{PubMedQA}: {A} Dataset for Biomedical Research Question Answering}},
  author={Jin, Qiao and Dhingra, Bhuwan and Liu, Zhiyong and 
  Cohen, William and Lu, Xinghua},
  booktitle={EMNLP},
  year={2019}
}

@article{abdin2024phi3,
  title={{Phi-3 Technical Report: {A} Highly Capable Language Model 
  Locally on Your Phone}},
  author={Abdin, Marah and others},
  journal={arXiv preprint arXiv:2404.14219},
  year={2024}
}

@article{grattafiori2024llama,
  title={{The Llama 3 Herd of Models}},
  author={Grattafiori, Aaron and others},
  journal={arXiv preprint arXiv:2407.21783},
  year={2024}
}

@article{team2024gemma,
  title={{Gemma 2: Improving Open Language Models at a Practical Size}},
  author={{Gemma Team} and others},
  journal={arXiv preprint arXiv:2408.00118},
  year={2024}
}

@article{jiang2023mistral,
  title={{Mistral} {7B}},
  author={Jiang, Albert Q and others},
  journal={arXiv preprint arXiv:2310.06825},
  year={2023}
}

@article{bedi2025systematic,
  title={{A Systematic Review of Large Language Model Evaluations 
  in Clinical Medicine}},
  author={Bedi, Sehej and others},
  journal={BMC Medical Informatics and Decision Making},
  year={2025}
}

@article{kim2025hallucination,
  title={{Medical Hallucinations in Foundation Models and Their 
  Impact on Healthcare}},
  author={Kim, Yubin and others},
  journal={arXiv preprint},
  year={2025}
}

@article{white2023prompt,
  title={{A Prompt Pattern Catalog to Enhance Prompt Engineering 
  with ChatGPT}},
  author={White, Jules and others},
  journal={arXiv preprint arXiv:2302.11382},
  year={2023}
}

@article{ngweta2024robustness,
  title={{Towards {LLMs} Robustness to Changes in Prompt Format Styles}},
  author={Ngweta, Lilian and others},
  journal={arXiv preprint},
  year={2024}
}

@article{lewis2020rag,
  title={{Retrieval-Augmented Generation for 
  Knowledge-Intensive {NLP} Tasks}},
  author={Lewis, Patrick and others},
  journal={NeurIPS},
  year={2020}
}

@article{cleverjmir2025,
  title={{Clinical Large Language Model Evaluation by Expert 
  Review ({CLEVER}): Framework Development and Validation}},
  author={Kocaman, Veysel and others},
  journal={JMIR AI},
  year={2025}
}

@article{garg2024slm,
  title={{A Survey of Small Language Models in Healthcare}},
  author={Garg, Sehej and others},
  journal={arXiv preprint},
  year={2024}
}

@article{chen2023meditron,
  title={{{MEDITRON-70B}: Scaling Medical Pretraining for 
  Large Language Models}},
  author={Chen, Zeming and others},
  journal={arXiv preprint arXiv:2311.16079},
  year={2023}
}

@misc{github2026,
  title={Clinical {LLM} Consistency Study -- Code and Data},
  author={Hariprasad, Shravani},
  year={2026},
  howpublished={\url{https://github.com/shravani-01/clinical-llm-eval}}
}

\appendix

\section{Representative Prompt Variation Example}
\label{appendix:prompt_example}

This appendix presents a single representative example illustrating
how the five prompt styles were applied to an identical clinical
question, along with the raw model responses and extracted answers
from Phi-3 Mini. This example was selected because it demonstrates
prompt sensitivity in practice: the model produces the correct
answer under three prompt styles (Formal, Simplified, Direct) but
selects an incorrect answer under both the Original and Roleplay
styles, despite identical question content across all five
variations.

\subsection*{Question}

A healthy 23-year-old male is undergoing an exercise stress test
as part of his physiology class. If blood were to be sampled at
different locations before and after the stress test, which area
of the body would contain the lowest oxygen content at both time
points?

\medskip
\noindent\textbf{Options:}
\begin{itemize}
  \item[A:] Inferior vena cava
  \item[B:] Coronary sinus \textit{(correct answer)}
  \item[C:] Pulmonary artery
  \item[D:] Pulmonary vein
\end{itemize}

\subsection*{Prompt Texts and Model Responses (Phi-3 Mini)}

\begin{table}[H]
\begin{singlespace}
\centering
\small
\caption{Prompt texts, raw model output, and extracted answers for
Phi-3 Mini on a representative MedQA question. The correct answer
is B (Coronary sinus). The Roleplay style produces an incorrect
answer (A) despite identical question content across all five
variations.}
\label{tab:appendix_example}
\resizebox{\textwidth}{!}{%
\begin{tabular}{p{1.8cm} p{7.5cm} p{3.5cm} >{\centering\arraybackslash}p{1.8cm} >{\centering\arraybackslash}p{1.8cm}}
\toprule
\textbf{Style} & \textbf{Prompt Text (abbreviated)} & \textbf{Raw Response} & \textbf{Extracted} & \textbf{Correct?} \\
\midrule
Original &
\textit{``Answer the following medical question by choosing the correct option. \ldots Answer with only the option letter (A, B, C, or D).''} &
\texttt{Option A: Inferior vena c\ldots} &
A & No \\
\midrule
Formal &
\textit{``You are a medical expert. Based on your clinical knowledge, select the most appropriate answer \ldots Respond with only the letter of the correct choice (A, B, C, or D).''} &
\texttt{B: Coronary sinus} &
B & Yes \\
\midrule
Simplified &
\textit{``Read this medical question carefully and pick the best answer. \ldots Which letter is correct? Reply with just A, B, C, or D.''} &
\texttt{B: Coronary sinus The\ldots} &
B & Yes \\
\midrule
Roleplay &
\textit{``You are a physician taking a medical licensing exam. Answer this question as you would on the exam. \ldots Your answer (A, B, C, or D):''} &
\texttt{The correct answer is A: Inferior v\ldots} &
A & No \\
\midrule
Direct &
\textit{``Medical question: \ldots What is the correct option? State only the letter.''} &
\texttt{B} &
B & Yes \\
\bottomrule
\end{tabular}}
\end{singlespace}
\end{table}

\noindent The majority answer across five styles is B (3 of 5
responses), yielding a consistency score of $3/5 = 0.60$ and a
correct majority answer. However, the Roleplay prompt produces
answer A in both instances it appears, illustrating how persona-based
framing systematically shifts the model toward an incorrect response.
The Original prompt also returns A, despite minimal structural
difference from the Direct prompt --- a further illustration of
prompt sensitivity even between superficially similar formulations.
The raw responses also reflect a secondary finding: the 10-token
output limit causes truncation (\texttt{Option A: Inferior vena
c\ldots}), contributing to the UNKNOWN rate observed in cases where
truncation prevents a valid letter from being extracted.

\medskip
\noindent Full prompt texts for all five styles across all three
datasets are available in the project repository \citep{github2026}.

\subsection*{Example 2: Reliable Incorrectness (Gemma 2, Consistency = 1.0, Accuracy = 0)}

This example illustrates the \textit{reliable incorrectness} failure
mode described in Section~5.1. Gemma 2 selects the same incorrect
answer (D) across all five prompt styles, yielding a perfect
consistency score of 1.0 while being entirely wrong. The correct
answer is B (Isosorbide dinitrate), consistent with the presentation
of Prinzmetal angina, which is treated with nitrates and calcium
channel blockers rather than beta-blockers such as propranolol (D).
This case demonstrates that a model appearing maximally stable under
prompt variation can simultaneously be maximally unreliable from a
clinical standpoint.

\subsection*{Question}

A 38-year-old woman presents to her physician's clinic for recurrent
episodes of chest pain that wakes her from her sleep. While usually
occurring late at night, she has also had similar pains during the
day at random times, most recently while sitting at her desk in her
office and at other times while doing the dishes at home. The pain
lasts 10--15 minutes and resolves spontaneously. She is unable to
identify any common preceding event to pain onset. The remainder of
her history is unremarkable and she takes no regular medications.
Examination reveals: pulse 70/min, respirations 16/min, and blood
pressure 120/70 mmHg. A physical examination is unremarkable. Which
of the following would be effective in reducing her symptoms?

\medskip
\noindent\textbf{Options:}
\begin{itemize}
  \item[A:] Aspirin
  \item[B:] Isosorbide dinitrate \textit{(correct answer)}
  \item[C:] Heparin
  \item[D:] Propranolol
\end{itemize}

\subsection*{Model Responses (Gemma 2)}

\begin{table}[H]
\begin{singlespace}
\centering
\small
\caption{Gemma 2 responses across all five prompt styles for a
representative MedQA question demonstrating reliable incorrectness.
The correct answer is B (Isosorbide dinitrate). Gemma 2 selects D
(Propranolol) under every prompt style, yielding a consistency score
of 1.0 and 0\% accuracy on this question.}
\label{tab:appendix_reliable_incorrect}
\resizebox{\textwidth}{!}{%
\begin{tabular}{>{\raggedright\arraybackslash}p{1.8cm}
                >{\raggedright\arraybackslash}p{7.5cm}
                >{\raggedright\arraybackslash}p{3.5cm}
                >{\centering\arraybackslash}p{1.8cm}
                >{\centering\arraybackslash}p{1.8cm}}
\toprule
\textbf{Style} & \textbf{Prompt Text (abbreviated)} & \textbf{Raw Response} & \textbf{Extracted} & \textbf{Correct?} \\
\midrule
Original   & \textit{``Answer the following medical question by choosing the correct option. \ldots Answer with only the option letter (A, B, C, or D).''} & \texttt{D} & D & No \\
\midrule
Formal     & \textit{``You are a medical expert. Based on your clinical knowledge, select the most appropriate answer \ldots Respond with only the letter of the correct choice (A, B, C, or D).''} & \texttt{D} & D & No \\
\midrule
Simplified & \textit{``Read this medical question carefully and pick the best answer. \ldots Which letter is correct? Reply with just A, B, C, or D.''} & \texttt{D} & D & No \\
\midrule
Roleplay   & \textit{``You are a physician taking a medical licensing exam. Answer this question as you would on the exam. \ldots Your answer (A, B, C, or D):''} & \texttt{The correct answer is **(D) Propran\ldots} & D & No \\
\midrule
Direct     & \textit{``Medical question: \ldots What is the correct option? State only the letter.''} & \texttt{D} & D & No \\
\bottomrule
\end{tabular}}
\end{singlespace}
\end{table}

\noindent Consistency score = $5/5 = 1.0$; majority answer = D
(incorrect). This pattern --- perfect stability, zero correctness ---
exemplifies why consistency alone is insufficient as a clinical
safety metric. A clinician or automated system relying on answer
stability as a proxy for confidence would receive maximally
misleading signals from this model on this question. Gemma 2
produced 77 such cases on MedQA alone (out of 200 questions),
representing 38.5\% of its evaluation set.

\medskip
\noindent Full prompt texts for all five styles across all three
datasets are available in the project repository \citep{github2026}.

\end{document}